\documentclass[11pt]{article}

\usepackage{fullpage}
\usepackage[ruled, linesnumbered, vlined, commentsnumbered]{algorithm2e}
\usepackage{graphicx,subfig} 
\usepackage{amsfonts,amssymb,amsmath,amsthm,amsopn}	
\usepackage{hhline,booktabs,diagbox,colortbl,multirow,tabularx,threeparttable} 
\usepackage[listings,skins,breakable]{tcolorbox}

\usepackage{enumerate}
\usepackage{authblk}
\usepackage{footnote}
\usepackage{hyperref}
\usepackage{prettyref}
\usepackage{cite}
\usepackage{setspace}
\usepackage{color}
\usepackage{xcolor}  
\usepackage{scrpage2}
\usepackage{lscape}
\usepackage{geometry}

\usepackage{tikz}
\usepackage{pgfplots}
\usetikzlibrary{positioning,shapes,shadows,arrows,calc}
\tikzstyle{component}=[rectangle, draw=black, rounded corners, fill=blue!40, drop shadow, text centered, anchor=north, text=white, minimum height=1cm]
\tikzstyle{arrow}=[->, thick]

\pgfplotsset{compat=1.12}
\usetikzlibrary{intersections}
\usetikzlibrary{pgfplots.statistics}
\usepgfplotslibrary{fillbetween}

\definecolor{myblue}{RGB}{34,31,217}
\definecolor{mycyan}{gray}{.7}
\definecolor{Gray}{gray}{0.9}

\newcommand{\pref}{\prettyref}

\newrefformat{fig}{Fig.~\ref{#1}}
\newrefformat{tab}{Table~\ref{#1}}
\newrefformat{sec}{Section~\ref{#1}}
\newrefformat{app}{Appendix~\ref{#1}}
\newrefformat{alg}{Algorithm~\ref{#1}}
\newrefformat{property}{Property~\ref{#1}}
\newrefformat{theorem}{Theorem~\ref{#1}}
\newrefformat{corollary}{Corollary~\ref{#1}}
\newrefformat{proposition}{Proposition~\ref{#1}}
\newrefformat{def}{Definition~\ref{#1}}
\newrefformat{eq}{equation~(\ref{#1})}

\title{\vspace{-1ex}\LARGE\textbf{Adaptive Operator Selection Based on Dynamic Thompson Sampling for MOEA/D}\footnote{This manuscript is submitted for possible publication. Reviewer can use this version interchangeable.}}

\author[1]{\normalsize Lei Sun}
\author[2]{\normalsize Ke Li}
\affil[1]{\normalsize College of Computer Science and Engineering, University of Electronic Science and Technology of China, 611731, Chengdu, China}
\affil[2]{\normalsize Department of Computer Science, University of Exeter, EX4 4QF, Exeter, UK}
\affil[$\ast$]{\normalsize Email: \texttt{k.li@exeter.ac.uk}}

\date{}

\begin{document}
\maketitle

\vspace{-3ex}
{\normalsize\textbf{Abstract: }In evolutionary computation, different reproduction operators have various search dynamics. To strike a well balance between exploration and exploitation, it is attractive to have an adaptive operator selection (AOS) mechanism that automatically chooses the most appropriate operator on the fly according to the current status. This paper proposes a new AOS mechanism for multi-objective evolutionary algorithm based on decomposition (MOEA/D). More specifically, the AOS is formulated as a multi-armed bandit problem where the dynamic Thompson sampling (DYTS) is applied to adapt the bandit learning model, originally proposed with an assumption of a fixed award distribution, to a non-stationary setup. In particular, each arm of our bandit learning model represents a reproduction operator and is assigned with a prior reward distribution. The parameters of these reward distributions will be progressively updated according to the performance of its performance collected from the evolutionary process. When generating an offspring, an operator is chosen by sampling from those reward distribution according to the DYTS. Experimental results fully demonstrate the effectiveness and competitiveness of our proposed AOS mechanism compared with other four state-of-the-art MOEA/D variants.

}


\section{Introduction}
\label{sec:introduction}

Multi-objective optimisation problems (MOPs) are ubiquitous in various scientific~\cite{ClarkW96} and engineering domains~\cite{PierroKSB09}. Due to the conflicting nature of different objectives, there does not exist a global optimum that optimises all objectives simultaneously. Instead, multi-objective optimisation aims to find a set of trade-off alternatives, an improvement at one objective of which can lead to a degradation of at least one other objective, that approximate the Pareto-optimal front (PF).

Due to the population-based characteristics, evolutionary algorithm (EA) has been recognised as the major approach for multi-objective optimisation. Over the past three decades and beyond, a significant amount of efforts have been devoted to the development of evolutionary multi-objective optimisation (EMO) algorithms~\cite{LiZZL09,LiZLZL09,LiKWCR12,LiKCLZS12,LiKWTM13,LiK14,WuKZLWL15,LiKZD15,LiKD15,LiDZ15,LiDZZ17,WuKJLZ17,WuLKZZ17,LiDY18,ChenLY18,WuLKZ20,ChenLBY18,LiCFY19,WuLKZZ19,LiCSY19,Li19,ZouJYZZL19}, e.g., fast non-dominated sorting genetic algorithm II (NSGA-II) \cite{DebAPM02}, indicator-based EA (IBEA) \cite{ZitzlerK04} and multi-objective EA based on decomposition (MOEA/D) \cite{ZhangL07}. It is widely appreciated that the search behaviour of an EA largely depends on its reproduction operator(s). In particular, some operators are exploration-oriented and are good at exploring unknown regions in the search space; whilst the others are exploitation-oriented and mainly focus on exploiting the current superior regions. How to strike a balance between exploration and exploitation is a long standing topic in order to achieve an efficient and effective evolutionary search process.

Adaptive operator selection (AOS) is an emerging paradigm that aims to autonomously select the most appropriate reproduction operator according to the latest search dynamics. Generally speaking, an AOS paradigm consists of two major steps~\cite{LiFK11,LiWKC13}. One is credit assignment that gives an operator a reward according to its up to date performance; the other is decision-making that selects the `appropriate' operator for the next stage according to the accumulated awards. A fundamental issue behind the AOS is an exploration versus exploitation (EvE) dilemma. One hopes to give more chances to operators with decent track records (exploitation), but also wants to explore poor operators in the future search (exploration), since an operator might perform significantly differently at different search stages. It is worth noting that credit assignment under a multi-objective setting is even more challenging due to the conflicting characteristics of different objectives. To address the EvE dilemma, Li et al. \cite{LiFKZ14} initially transformed the AOS problem into a multi-armed bandits (MAB) problem and applied the classic upper confidence bound (UCB) algorithm \cite{Auer02} to implement an AOS paradigm in EMO. As for the difficulty of credit assignment in multi-objective optimisation, MOEA/D is used as the baseline given that it decomposes the original MOP into several single objective subproblems which facilitate the fitness evaluation.

However, one of the major concerns of using bandit learning for AOS is its stationary environment assumption used in the traditional MAB model. In other words, the probability distribution of the reward for pulling an arm is fixed a \textit{priori} whereas the evolutionary search process is highly non-stationary. Bearing this consideration in mind, this paper proposes to use the dynamic Thompson sampling strategy \cite{GuptaGA11} to address the EvE dilemma under a non-stationary environment. More specifically, in our AOS paradigm, each reproduction operator is regarded as an arm in a bandit game and is assigned with a prior reward distribution. During the evolutionary search process, the credit of each operator is updated according to the fitness improvement achieved by using the corresponding operator along with the parameters associated with its reward distribution. During the decision-making step, a reproduction operator is selected by sampling from those reward distributions. To facilitate the fitness evaluation under a multi-objective setting, we carry on with the MOEA/D as the baseline and the end algorithm is denoted as MOEA/D-DYTS. From our experiments, we have witnessed the superior performance of MOEA/D-DYTS over other four state-of-the-art MOEA/D variants on 19 benchmark test problems.

The remainder of this paper is organised as follows. \pref{sec:background} provides some preliminary knowledge. \pref{sec:method} delineates the technical details of our proposed MOEA/D-DYTS step by step. The performance of MOEA/D-DYTS is validated in~\pref{sec:experiment}. At the end, \pref{sec:conclusion} concludes this paper and shed some lights on future directions.


\section{Preliminaries}
\label{sec:background}

In this section, we provide the preliminary knowledge, including some definitions related to multi-objective optimisation and the baseline algorithm, required in this paper.
\subsection{Multiobejctive Optimization Problems}
Without loss of generality, the MOP considered in this paper is defined as:
\begin{equation}
\begin{aligned}
    &\mathrm{minimize} \quad \mathbf{F}(\mathbf{x})=\left(f_1(\mathbf{x}),\cdots,f_m(\mathbf{x})\right)\\
    &\mathrm{subject\ to } \,\,\, \mathbf{x}\in\mathrm{\Omega}
\end{aligned},
\end{equation}
where $\mathbf{x} = (x_1,\cdots,x_n) \in\mathrm{\Omega}$ is a decision variable vector, $\mathrm{\Omega}=\mathrm{\Pi}_{i=1}^n[ l_i,u_i]\in\mathbb{R}^n$ is the decision space where $l_i$ and $u_i$ are the lower and upper bounds of the $i$-th dimension. $\mathbf{F}:\mathrm{\Omega}\rightarrow\mathbb{R}^m$ consists of $m$ conflicting objective functions. Given two decision vectors $\mathbf{x}_1$ and $\mathbf{x}_2\in\mathrm{\Omega}$, $\mathbf{x}_1$ is said to dominate $\mathbf{x}_2$, denoted as $\mathbf{x}_1\preceq \mathbf{x}_2$, if and only if $f_i(x_1)\le f_i(x_2)$ for all $i\in \{1,\cdots,m\}$ and $\mathbf{F}(\mathbf{x}_1) \neq \mathbf{F}(\mathbf{x}_2)$. A solution $\mathbf{x^\ast}\in\mathrm{\Omega}$ is said to be Pareto-optimal when no other solution $\mathbf{x}\in\mathrm{\Omega}$ can dominate $\mathbf{x^\ast}$. The set of all Pareto-optimal solutions is called Pareto-optimal set (PS) whilst its image in the objective space is called the PF.

\subsection{Baseline Algorithm}
In this paper, we use the MOEA/D as the baseline EMO framework. The basic idea of MOEA/D is to decompose the original MOP into several subproblems, each of which is either as an aggregated scalarising function or simplified MOP. Thereafter, MOEA/D uses a population-based meta-heuristic to solve these subproblems in a collaborative manner. In particular, the widely used Tchebychff function is used to form the subproblem in this paper and it is defined as: 
\begin{equation}
\min _{\mathbf{x} \in \mathrm{\Omega}} g^{\text {tch }}\left(\mathbf{x} |\mathbf{w},\mathbf{z^{*}}\right)=\max _{1 \leq i \leq m}\left\{\left|f_{i}(\mathbf{x})-\mathbf{z}_{i}^{*}\right| / w_{i}\right\},
\end{equation}
where $\mathbf{w}=(w_1,\cdots,w_m)$ is a weight vector, evenly sampled from a canonical simplex, with $w_i \geq 0$, $i \in \{ 1,\cdots,m\}$, and $\sum_{i=1}^{m} w_{i}=1$. $\mathbf{z}^{*}=(z_{1}, \cdots, z_{m})$ is the ideal objective where $z_{i}=\min_{\mathbf{x} \in PS}f_{i}(x) $, $i\in \{1,\cdots,m\}$. From the above description, we can see that the quality of a solution $\mathbf{x}\in \mathrm{\Omega}$ can be evaluated by a subproblem $g^{\text {tch }}\left(\mathbf{x} |\mathbf{w},\mathbf{z}^{*}\right)$ which facilitates the credit assignment under a multi-objective setting.

Note that instead of using the vanilla MOEA/D, here we use its variant with a dynamic resource allocation scheme, dubbed as MOEA/D-DRA \cite{ZhangLL09}, as the baseline algorithm given its outstanding performance in CEC 2009 MOEA competition. Different from the vanilla MOEA/D where all subproblems are allocated with the same amount of the computational resources, MOEA/D-DRA dynamically selects some most promising subproblems to evolve according to their utilities defined as:
\begin{equation}
\label{uti_cal1}
\pi^{i}=\left\{\begin{array}{ll}
1 & \text { if } \Delta^{i}>0.001 \\
\left(0.95+0.05 \times \frac{\Delta^{i}}{0.001}\right) \times \pi^{i} & \text { otherwise }
\end{array}\right.,
\end{equation}
where $\Delta^i$ is the fitness improvement rates (FIR) of the objective function value in subproblem $i$, which is defined as:
\begin{equation}
\label{uti_cal2}
\Delta^{i}=\frac{g^{\text{tch }}(\mathbf{x}_{t-\Delta t}^{i} | \mathbf{w}^{i}, \mathbf{z}^{*})-g^{\text {tch }}\left(\mathbf{x}_{t}^{i} | \mathbf{w}^{i}, \mathbf{z}^{*}\right)}{g^{\text {tch }}\left(\mathbf{x}_{t-\Delta t}^{i} | \mathbf{w}^i, \mathbf{z}^{*}\right)},
\end{equation}
where $t$ is the current generation, and $\Delta t$ is the updating period. Interested readers are referred to~\cite{ZhangLL09} for more details of MOEA/D-DRA.


\section{Proposed Algorithm }
\label{sec:method}

In this section, we delineate the implementation our proposed MOEA/D-DYTS. Specifically, we will start with the classical Bernoulli MAB problem and the vanilla Thompson sampling. Thereafter, we develop the definition of the dynamic Thompson sampling strategy which is the main crux of our AOS paradigm. At the end, we give the algorithmic implementation of MOEA/D-DYTS.

\subsection{Thompson Sampling}
\label{sec:ts}
In probability theory and machine learning, the MAB problem considers optimally allocating a fixed set of limited resources among competing actions $\mathcal{A} = \{a_1,\cdots,a_k\}$ that finally maximises the expected return (or gain). In this paper, we consider a traditional Bernoulli MAB problem where the reward of conducting an action $a_i$ is $r_i\in\{0,1\},i\in\{1,\cdots,k\}$. At the time step $t$, the probability of obtaining a reward of one by conducting the action $a_i$ is $\mathbb{P}(r_i = 1|a_i,\theta)=\theta_i$. On the other hand, $\mathbb{P}(r_i = 0|a_i,\theta)=1-\theta_i$ when obtaining a reward of none. In particular, $\theta_i$ is defined as the mean reward of action $a_i,i\in\{1,\cdots,k\}$. Under the MAB setting, the mean rewards for all actions $\mathrm{\theta} = (\theta_1,\cdots,\theta_k)$ are fixed over time but are unknown beforehand. The ultimate goal of this Bernoulli MAB problem is to maximise the cumulative rewards over a period of $T$ time steps.

\begin{figure}[h]
\centering
\includegraphics[width=6cm]{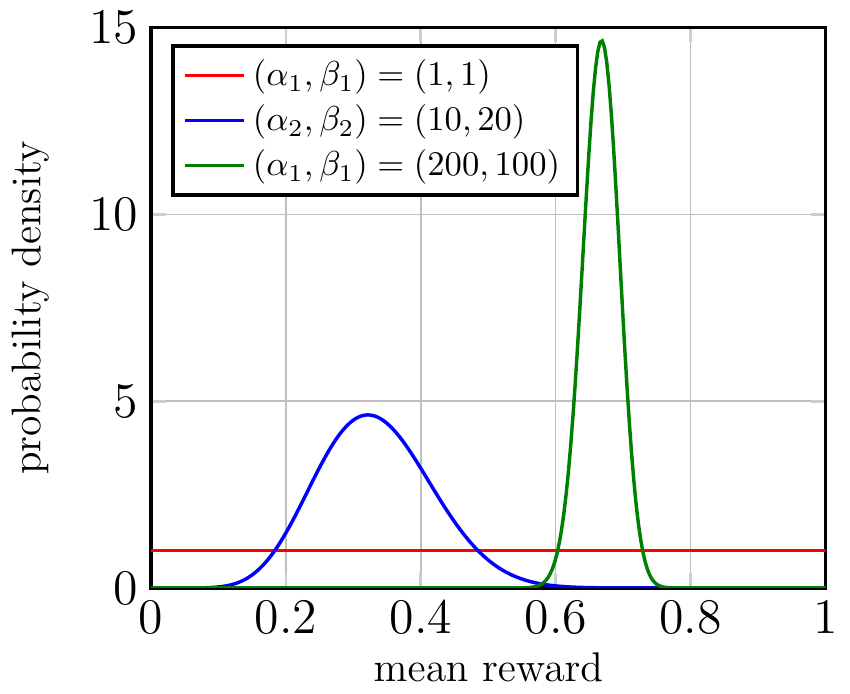}
\caption{Illustration of Beta distribution with different $(\alpha,\beta)$ settings.}
\label{fig_beta}
\end{figure}

Thompson sampling \cite{Thompson33} is an efficient algorithm for solving the Bernoulli MAB problem. It takes advantage of Bayesian estimation for online decision problems where actions are applied in a sequential manner. In Thompson sampling, each arm is assigned with an independent prior reward distribution of $\theta_i,i\in\{1,\cdots,k\}$. In particular, we use the Beta distribution with parameters $\alpha = \left(\alpha_{1},\dots,\alpha_{k}\right)$ and $\beta = \left(\beta_{1},\dots,\beta_{k}\right)$ to represent this prior distribution:
\begin{equation}
\mathcal{P}^{Beta}\left(\theta_{i}\right)=\frac{\Gamma\left(\alpha_{i}+\beta_{i}\right)}{\Gamma\left(\alpha_{i}\right) \Gamma\left(\beta_{i}\right)} \theta_{i}^{\alpha_{i}-1}\left(1-\theta_{i}\right)^{\beta_{i}-1},
\end{equation}
where $\alpha_i$ and $\beta_i$ are the parameters associated with the $i$-th arm, $i\in\{1,\cdots,k\}$ and $\Gamma(\cdot)$ is the gamma function. The parameters of this distribution are updated according to the Bayes' rule after receiving the up to date rewards. Due to the conjugate properties between Beta distribution and Bernoulli distribution, the posterior distribution of each action is still a Beta distribution. For a given action $a_i,i\in\{1,\cdots,k\}$, if the reward of its application is $r_i = 1$, then $\alpha_i$ and $\beta_i$ are updated as:
\begin{equation}
\alpha_i = \alpha_i + 1,\beta_i = \beta_i.  
\end{equation}
Otherwise, if $r_i = 0$ then we have:
\begin{equation}
\alpha_i = \alpha_i,\beta_i = \beta_i + 1,
\end{equation}
whilst the other parameters are kept unchanged. In the decision-making stage, the arm having the largest sampling value from the distribution is chosen for the next step. This is the underlying mechanism of Thompson sampling for balancing exploration and exploitation. In practice, each component of $\alpha$ and $\beta$ is initialised to be 1 whilst $\mathcal{P}^{Beta}(\theta_i)$ is set to be a uniform distribution over $[0,1]$. The Beta distribution has a mean $\alpha_i/(\alpha_i+\beta_i)$ and its distribution become more concentrated at its mean as $\alpha_i+\beta_i$ grows. For example, as shown in Fig.~\ref{fig_beta}, the probability density function of $(\alpha_2,\beta_2)=(200,100)$ is the most concentrated one whilst that of $(\alpha_1,\beta_1)=(1,1)$ is a flat line.
In order to apply Thompson sampling for AOS in an EA, we consider each reproduction operator as a bandit arm with a Bernoulli distribution. In particular, an operator receives a reward of one in case its application improves the evolutionary population towards optimum, otherwise its reward is none.

\subsection{Dynamic Thompson Sampling}
\label{sec:dyts}
Since the reward distribution used in the aforementioned vanilla Thompson sampling is fixed \textit{a priori}, it does not fit the non-stationary nature of EAs. In other words, the reward distribution of a reproduction operator is highly likely to change with the progress of the evolution. To address this issue, this paper considers using a dynamic Thompson sampling \cite{GuptaGA11} strategy (DYTS) to serve as the foundation of our AOS paradigm. In particular, DYTS mainly uses a threshold $C$ to control the two different parameter update rules. If $\alpha_i + \beta_i < C$, the parameters of the Beta distribution are updated as the vanilla Thompson sampling introduced in~\pref{sec:ts}. On the other hand, if $\alpha_i + \beta_i \geq C$, the corresponding update rule aims to tweak the parameters so that we have $\alpha_i + \beta_i = C$ after this update. The pseudocode of the parameter update strategy of DYTS is given in~\pref{alg:update_rule}.

\begin{algorithm}[t]
\scriptsize
\caption{\texttt{ParameterUpdate}$(\alpha_i,\beta_i,r,C)$: parameter update rule of the DYTS strategy}
\label{alg:update_rule}
\KwIn{Parameters of the $i$-th operator $\alpha_i$, $\beta_i$, its up to date reward $reward$ and threshold $C$}
\KwOut{Updated $(\alpha_i,\beta_i)$}
\eIf{$\alpha_i + \beta_i < C$}{
$\alpha_i = \alpha_i + r$\;
$\beta_i = \beta_i + 1 - r$\;
}{
$\alpha_i = (\alpha_i + r)\frac{C}{C+1}$\;
$\beta_i = (\beta_i + 1 - r)\frac{C}{C+1}$\;
}
\Return $(\alpha_i,\beta_i)$\;
\end{algorithm}

There are two characteristics of this parameter update rule.
\begin{itemize}
	\item It ensures that $\alpha_i+\beta_i\leq C$. More specifically, if $\alpha_i+\beta_i=C$, we have 
	\begin{equation}
		\begin{aligned}
			\alpha^\prime_{i} + \beta^\prime_{i} &= (\alpha_i + \beta_i + 1)\frac{C}{C+1}\\
			&=(C+1)\frac{C}{C+1}\\
			&=C
		\end{aligned},
	\end{equation}
	where $\alpha^\prime_{i}$ and $\beta^\prime_{i}$ represent the updated $\alpha_i$ and $\beta_i$. In this case, we are able to avoid concentration of samples coming out of the DYTS strategy. Henceforth, it enables the DYTS a certain level of exploration ability.
	\item It gives larger weights to those recent awards thus help track the up to date reward distribution. More specifically, let us denote the reward received by the $i$-th reproduction operator at the $t$-th update as $r_i^t$, the corresponding parameter $\alpha_i^t$ can be derived as:
\begin{equation}
\begin{aligned}
\alpha^t_{i}
&=(\alpha^{t-1}_i + r^t_i)\frac{C}{C+1}\\
&=((\alpha^{t-2}_i + r^{t-1}_i)\frac{C}{C+1} + r^t_i)\frac{C}{C+1}\\
&=\alpha^{t-2}_i (\frac{C}{C+1})^2 + r^{t-1}_i(\frac{C}{C+1})^2 + r^t_i(\frac{C}{C+1}).
\end{aligned}
\end{equation}
Analogously, we can have the same derivation of $\beta^t_i$. According to this derivation, we can see that the latest reward is assigned with the largest decay ratio. By this means, the DYTS strategy enables the Thompson sampling to a non-stationary environment.
\end{itemize}

\subsection{Operator Pool}
\label{sec:operators}

From our preliminary experiments, this paper considers using the following five reproduction operator to constitute our operator pool. In particular, four of them are differential evolution (DE) variants directly derived from our previous paper~\cite{LiFKZ14} whilst the other one is the uniform mutation (UM)~\cite{Deb01}.
\begin{itemize}
\item DE/rand/1: $\mathbf{x}^{c}=\mathbf{x}^{i}+F *\left(\mathbf{x}^{r_{1}}-\mathbf{x}^{r_{2}}\right)$,
\item DE/rand/2: $\mathbf{x}^{c}=\mathbf{x}^{i}+F *\left(\mathbf{x}^{r_{1}}-\mathbf{x}^{r_{2}}\right)+F *\left(\mathbf{x}^{r_{3}}-\mathbf{x}^{r_{4}}\right)$,
\item DE/current-to-rand/1: $\mathbf{x}^{c}=\mathbf{x}^{i}+K *\left(\mathbf{x}^{i}-\mathbf{x}^{r_{1}}\right)+F *\left(\mathbf{x}^{r_{2}}-\mathbf{x}^{r_{3}}\right)$,
\item DE/current-to-rand/2: $\mathbf{x}^{c}=\mathbf{x}^{i}$+$K *\left(\mathbf{x}^{i}-\mathbf{x}^{r_{1}}\right)$+$F *\left(\mathbf{x}^{r_{2}}-\mathbf{x}^{r_{3}}\right)$+$F * \left(\mathbf{x}^{r_{4}}-\mathbf{x}^{r_{5}}\right)$,
\item UM: $\mathbf{x}^{c}=\mathbf{x}^{i}+ \texttt{uniform}(0,1)*(\mathbf{u}-\mathbf{l})$,
\end{itemize}
where $\texttt{uniform(0,1)}$ represents a uniform distribution within the range [0,1], $\mathbf{u}=(u_1,\cdots,u_n)$ and $\mathbf{l} = (l_1,\cdots,l_n)$ indicate the upper and lower bounds of decision space.

\subsection{Credit Assignment}
\label{sec:credit}

It is worth noting that the reward considered in the Bernoulli distribution is a binary value thus is not directly applicable in an EA. To address this issue, this paper uses the fitness improvement (FI) as the measure to evaluate the credit of an application of the reproduction operator chosen by the AOS. Formally, FI is defined as:
\begin{equation}
FI = \max_{\mathbf{x}^i\in P}\{g^{\text{tch }}(\mathbf{x}^{i} | \mathbf{w}^{i}, \mathbf{z}^{*})-g^{\text {tch }}\left(\mathbf{x}^{c} | \mathbf{w}^{i}, \mathbf{z}^{*}\right)\},
\end{equation}
where $\mathbf{w}^{i}$ is the weight vector associated with the subproblem of a solution $\mathbf{x}^i\in P,i\in\{1,\cdots,N\}$ and $\mathbf{x}^{c}$ is the offspring solution generated by using the selected operator. If $FI>0$, it indicates that the application of the selected operator is successful thus the corresponding reward is one. Otherwise, the reward is set to be none instead.

\begin{algorithm}[t]
\scriptsize
\caption{\texttt{AOS}($\alpha,\beta$): adaptive operator selection based on DYTS}
\label{alg:AOS}
\KwIn{Parameters of the Beta distributions $\alpha=(\alpha_1,\cdots,\alpha_k)$ and $\beta=(\beta_1,\cdots,\beta_k)$}
\KwOut{Index of the selected reproduction operator}
\For{$i\leftarrow1$ \KwTo $k$}{
	Sample the estimated mean reward $\hat{\theta}_i$ from Beta distribution $\mathcal{P}^{Beta}(\alpha_i,\beta_i)$.
}
$a = \text{argmax}_{i\in\{1,\cdots,k\}}\hat{\theta}_i $\;
\Return $a$\;
\end{algorithm} 

\subsection{AOS Based on DYTS Strategy}
\label{sec:AOS}

The idea of our proposed AOS paradigm based on the DYTS strategy is simple and intuitive. As shown in~\pref{alg:AOS}, the operator having the largest sampling value from the up to date Beta distribution is chosen as the target operator for the next iteration.

\begin{algorithm}[t]
\scriptsize
\caption{MOEA/D-DYTS}
\label{alg:moead-dyts}
\KwIn{Algorithm parameters}
\KwOut{Approximated solution set $\mathcal{P}$}
Initialise the population $\mathcal{P}=\{\mathbf{x}^1,\cdots,\mathbf{x}^N\}$, the weight vectors $\{\mathbf{w}^1,\cdots,\mathbf{w}^N\}$, parameters of the Beta distribution $\alpha,\beta$ and the ideal point $\mathbf{z}^*$\;
$gen \leftarrow 0$, $neval \leftarrow 0$\;
\For{$i\leftarrow1$ \KwTo $N$}{
	$\mathcal{B}(i)\leftarrow\{i_1,\cdots,i_T\}$ where $\mathbf{w}^{i_1},\cdots,\mathbf{w}^{i_T}$ are the $T$ closest weight vectors to $\mathbf{w}^{i}$ and set $\pi_i \leftarrow1$\;
}
\While{$neval<maxEvaluations$}{
Let all the indices of the subproblems whose objectives are MOP individual objectives $f_i$ form the initial $I$. By using 10-tournament selection based on $\pi_i$, select other $\lfloor N / 5\rfloor- m$ indices and add them to $I$\;
\For{$each \, i \in I$}
{
	$op\leftarrow$ \texttt{AOS}$(\alpha,\beta)$\;
	\eIf{$uniform(0,1) < \delta$}
	{
		$P\leftarrow B(i)$\;
	}
	{
		$P\leftarrow$ the entire population \;
	}
	Randomly select a required number of parent solutions from $P$\;
	Generate an offspring $\mathbf{x}^c$ by the $op$-th operator over the selected solutions\;
	Use polynomial mutation to further mutate $\mathbf{x}^c$\;
	Update the ideal point $\mathbf{z}^*$ according to $\mathbf{x}^c$\;
	$FI \leftarrow \max_{\mathbf{x}^i\in P}\{g^{\text{tch }}(\mathbf{x}^{i} | \mathbf{w}^{i}, \mathbf{z}^{*})-g^{\text {tch }}\left(\mathbf{x}^{c} | \mathbf{w}^{i}, \mathbf{z}^{*}\right)\}$\;
	\eIf{$FI > 0$}
	{
		$reward \leftarrow 1$\;
		Replace the $\mathbf{x}^i$ associated with $FI$ by $\mathbf{x}^c$;
	}
	{
		$reward \leftarrow 0$\;
	}
	$(\alpha_{op},\beta_{op})\leftarrow\texttt{ParameterUpdate}(\alpha_{op},\beta_{op},reward,C)$\;
	$neval \leftarrow neval+1$\;

}
$gen\leftarrow gen+1$\;
\If{$modulo(gen,50) == 0$}
{
	Update the utility $\pi^i$ of each subproblem $i,i\in\{1,\cdots,N\}$;
}
}
\Return{P}\;
\end{algorithm}
\vspace{-0.25cm}

\subsection{Framework of MOEA/D-DYTS}
Our proposed AOS based on DYTS can be applied to MOEA/D-DRA (dubbed MOEA/D-DYTS) in a plug-in manner without any significant modification. In particular, we only need to maintain an operator pool and keep a record of the FI achieved by the application of an operator. As the pseudocode of MOEA/D-DYTS given in~\pref{alg:moead-dyts}, we can see that most parts are the same as the original MOEA/D-DRA. The only difference lies in the offspring reproduction where the AOS based on DYTS strategy is applied to select the most appropriate operator in practice (line 8 of~\pref{alg:moead-dyts}). In addition, after the generation of an offspring, the corresponding $FI$ is calculated followed by an update of the reward (lines 17 to 22 of~\pref{alg:moead-dyts}). Thereafter, the collected reward is used to update the parameters of the Beta distribution (line 23 of~\pref{alg:moead-dyts}).

\section{Experimental Studies}
\label{sec:experiment}

In this section, we will use a set of experiments to validate the effectiveness of our proposed MOEA/D-DYTS. The experimental settings used in this paper are briefly overviewed in~\pref{sec:experiment_set} including the benchmark test problems, parameter settings and the peer algorithms used in our experiments.

\subsection{Experimental Settings}
\label{sec:experiment_set}
In our experimental studies, 19 unconstrained test problems are used to constitute the benchmark suite including UF1 to UF10 from the CEC 2009 MOEA competition \cite{ZhangZZPLS08} and WFG1 to WFG9 chosen from the Walking Fish Group test problem set \cite{HubandHBW06}. In particular, the number of decision variables of UF problem is 30 whilst it is set to 38 (18 are position related and 20 are distance related) for the WFG problems~\cite{LiuLC19,LiXT19,GaoNL19}. Four state-of-the-art MOEA/D variants i.e., MOEA/D-FRRMAB \cite{LiFKZ14}, MOEA/D-GRA \cite{ZhouZ16}, MOEA/D-IRA \cite{LinJMWCLCZ18} and MOEA/D-DE \cite{LiZ09} are used as the peer algorithms in comparison. The parameters associated with these peer algorithms are set the same as recommended in their original paper. Those of our proposed MOEA/D-DYTS are set as follows:
\begin{itemize}
    \item The population size $N$ is set to 300 for the two-objective UF instances and 600 for the three-objective UF instances. As for WFG instances, we set $N=100$.
    \item Each algorithm is run 31 times on each test problem instance. The maximum number of function evaluations is set to $300,000$ for the UF instances and $25,000$ for the WFG instances. 
    \item The neighbourhood size is fixed to 20. Probability $\delta$ with regard to selecting $P$ is set to 0.8 as suggested in \cite{LinJMWCLCZ18}.
	\item The update threshold of our DYTS strategy is set as $C = 100$.
\end{itemize}
To evaluate the performance of different algorithms, two widely used performance metrics, i.e.,  inverted generational distance (IGD) \cite{BosmanT03} and Hypervolume (HV) \cite{ZitzlerT99}, are used in our experiments. Both of them are able to evaluate the convergence and diversity simultaneously. In order to calculate the IGD, 10,000 points were uniformly sampled from the true PF to constitute the reference set. The lower the IGD is, the better the solution set for approximating the PF. As for the HV calculation, we set the reference point as $(2.0,2.0)$ for two-objective UF instances and $(2.0,2.0,2.0)$ for three-objective UF instances. For the WFG instances, it is set as $(3.0,5.0)$. In contrast to the IGD, the larger the HV is, the better quality of the solution set for approximating the PF.

\begin{table}[h]

\scriptsize
  \centering
  \caption{Comparative results of all the algorithms on the UF and WFG test problems regarding IGD.
}
\vspace{-0.1cm}
\renewcommand{\arraystretch}{1.2}
\resizebox{\textwidth}{31mm}{
    \begin{tabular}{cccccc}
\toprule
          & MOEA/D-DE & MOEA/D-FRRMAB & MOEA/D-GRA & MOEA/D-IRA & MOEA/D-DYTS \\
\midrule
    UF1   & 1.97E-3$_{1.56E-4}$$^-$& 2.50E-3$_{2.19E-4}$$^-$&1.90E-3$_{8.3E-5}$$^\sim$ & 1.92E-3$_{8.9E-5}$$^\sim$ & \cellcolor[rgb]{ .682,  .667,  .667}\textbf{1.89E-3}$_\textbf{7.1E-5}$ \\
    UF2   & 7.12E-3$_{1.58E-3}$$^-$  & 5.64E-3$_{5.21E-4}$$^-$  & \cellcolor[rgb]{ .682,  .667,  .667}\textbf{3.92E-3}$_\textbf{5.01E-4}$$^\sim$  & 4.16E-3$_{5.92E-4}$$^\sim$  &4.09E-3$_{9.86E-4}$ \\
    UF3   & 1.20E-2$_{1.33E-2}$$^-$ & 6.97E-3$_{4.81E-3}$$^-$ & 4.24E-3$_{2.31E-3}$$^\sim$ & 5.27E-3$_{2.69E-3}$$^-$ & \cellcolor[rgb]{ .682,  .667,  .667}\textbf{4.04E-3}$_\textbf{2.07E-3}$ \\
    UF4   & 6.27E-2$_{4.09E-3}$$^-$ & 5.26E-2$_{3.88E-3}$$^-$ & 5.48E-2$_{3.41E-3}$$^-$ & 5.36E-2$_{2.85E-3}$$^-$ & \cellcolor[rgb]{ .682,  .667,  .667}\textbf{3.04E-2}$_\textbf{1.28E-3}$ \\
    UF5   & 3.12E-1$_{1.15E-1}$$^-$ & 2.90E-1$_{6.89E-2}$$^-$ & 2.41E-1$_{2.74E-2}$$^-$ & 2.35E-1$_{2.83E-2}$$^-$ & \cellcolor[rgb]{ .682,  .667,  .667}\textbf{1.26E-1}$_\textbf{2.37E-2}$\\
    UF6   & 1.85E-1$_{1.72E-1}$$^-$ & 2.07E-1$_{1.84E-2}$$^-$ & \cellcolor[rgb]{ .682,  .667,  .667}\textbf{7.43E-2}$_\textbf{2.97E-2}$$^+$ & 8.17E-2$_{5.06E-2}$$^+$ & 1.23E-1$_{6.38E-2}$ \\
    UF7   & 4.07E-3$_{4.24E-3}$$^-$ & 2.64E-3$_{3.37E-4}$$^-$ & 2.05E-3$_{1.02E-4}$$^-$ & 2.04E-3$_{1.03E-4}$$^-$ & \cellcolor[rgb]{ .682,  .667,  .667}\textbf{1.91E-3}$_\textbf{8.90E-5}$\\
    UF8   & 7.82E-2$_{1.27E-2}$$^-$ & 6.96E-2$_{1.29E-2}$$^\sim$ & 8.11E-2$_{1.43E-2}$$^-$ & 8.12E-2$_{1.20E-2}$$^-$ & \cellcolor[rgb]{ .682,  .667,  .667}\textbf{6.86E-2}$_\textbf{1.85E-2}$ \\
    UF9   & 8.82E-2$_{4.96E-2}$$^-$ & 7.97E-2$_{4.55E-2}$$^-$ & 3.82E-2$_{3.40E-2}$$^+$ & \cellcolor[rgb]{ .682,  .667,  .667}\textbf{3.49E-2}$_\textbf{2.79E-2}$$^+$ & 4.24E-2$_{2.55E-2}$ \\
    UF10  & 5.14E-1$_{6.85E-2}$$^-$ & 7.56E-1$_{1.19E-1}$$^-$ & 1.32E+00$_{2.39E-1}$$^-$ & 1.53E+00$_{3.24E-1}$$^-$ & \cellcolor[rgb]{ .682,  .667,  .667}\textbf{4.06E-1} $_\textbf{5.87E-2}$\\
    WFG1  & 1.28E+00$_{8.03E-3}$$^-$ & 1.28E+00$_{4.06E-3}$$^-$ & 1.26E+00$_{8.36E-3}$$^-$ & 1.26E+00$_{6.45E-3}$$^-$ & \cellcolor[rgb]{ .682,  .667,  .667}\textbf{1.20E+00}$_\textbf{6.08E-3}$ \\
    WFG2  & 4.06E-1$_{2.20E-1}$$^-$ & 2.54E-1$_{1.37E-1}$$^\sim$ & 2.91E-1$_{1.63E-1}$$^-$ & 2.59E-1$_{1.37E-1}$$^-$ & \cellcolor[rgb]{ .682,  .667,  .667}\textbf{2.33E-1}$_\textbf{1.22E-1}$ \\
    WFG3  & 1.55E-2$_{4.42E-3}$$^-$ & 1.40E-2$_{2.74E-3}$$^-$ & 1.39E-2$_{2.43E-3}$$^-$ & 1.39E-2$_{2.57E-3}$$^-$ & \cellcolor[rgb]{ .682,  .667,  .667}\textbf{1.32E-2}$_\textbf{1.00E-6}$ \\
    WFG4  & 2.89E-2$_{5.91E-3}$$^-$ & 2.81E-2$_{4.34E-3}$$^-$ & 1.87E-2$_{3.96E-3}$$^-$ & 2.07E-2$_{6.60E-3}$$^-$ & \cellcolor[rgb]{ .682,  .667,  .667}\textbf{1.72E-2}$_\textbf{1.89E-3}$ \\
    WFG5  & 4.23E-2$_{1.19E-2}$$^-$ & 3.88E-2$_{1.14E-2}$$^-$ & 3.02E-2$_{6.50E-3}$$^-$ & 3.06E-2$_{6.72E-3}$$^-$ & \cellcolor[rgb]{ .682,  .667,  .667}\textbf{2.12E-2}$_\textbf{1.78E-3}$\\
    WFG6  & 5.97E-2$_{2.89E-2}$$^-$ & 3.80E-2$_{2.30E-2}$$^-$ & 3.30E-2$_{9.99E-3}$$^-$ & 3.78E-2$_{1.00E-2}$$^-$ & \cellcolor[rgb]{ .682,  .667,  .667}\textbf{2.44E-2}$_\textbf{9.90E-3}$ \\
    WFG7  & 1.45E-2$_{4.89E-3}$$^-$ & 1.41E-2$_{3.76E-3}$$^-$ & 1.43E-2$_{4.41E-3}$$^-$ & 1.44E-2$_{4.36E-3}$$^-$ & \cellcolor[rgb]{ .682,  .667,  .667}\textbf{1.31E-2}$_\textbf{2.00E-6}$ \\
    WFG8  & 1.60E-2$_{6.14E-3}$$^-$ & 1.42E-2$_{4.10E-3}$$^-$ & 1.36E-2$_{2.63E-3}$$^\sim$ & 1.37E-2$_{3.30E-3}$$^\sim$ & \cellcolor[rgb]{ .682,  .667,  .667}\textbf{1.31E-2}$_\textbf{2.00E-6}$ \\
    WFG9  & 3.83E-2$_{5.54E-3}$$^-$ & 3.64E-2$_{5.11E-3}$$^-$ & 3.81E-2$_{7.61E-3}$$^-$ & 3.70E-2$_{4.96E-3}$$^-$ & \cellcolor[rgb]{ .682,  .667,  .667}\textbf{2.63E-2}$_\textbf{2.85E-3}$ \\
    -/$\sim$/+ & 19/0/0 & 17/2/0 & 13/4/2 & 14/3/2& \\
\bottomrule
    \end{tabular}}%
  \label{tab:IGD}%
\end{table}%

\vspace{-1em}

\subsection{Experimental Results}
In this section, we have compared MOEA/D-DYTS with four state-of-the-art MOEA/D variants, namely, MOEA/D-DE, MOEA/D-FRRMAB, MOEA/D-GRA and MOEA/D-IRA on UF instances and WFG instances. Tables \ref{tab:IGD} and \ref{tab:HV} present the result of the IGD and HV metric values obtained from 31 independent runs.  The best mean result for each problem is highlighted in boldface with grey background. Wilcoxon’s rank sum test with a 5\% significance level was also conducted to provide a statistically conclusion. Where ``-", ``+" and ``$\sim$" denote that the results obtained by corresponding algorithm are worse than, better than or similar to those of MOEA/D-DYTS.


\begin{table}[htbp]
\scriptsize
  \centering
  \caption{Comparative results of all the algorithms on the UF and WFG test problems regarding HV.
}
\vspace{-0.1cm}
\renewcommand{\arraystretch}{1.2}
\resizebox{\textwidth}{31mm}{
    \begin{tabular}{cccccc}
    \toprule
          & MOEA/D-DE & MOEA/D-FRRMAB & MOEA/D-GRA & MOEA/D-IRA & MOEA/D-DYTS \\
    \midrule
    UF1   & 3.65874$_{1.54E-3}$$^-$  & 3.65673$_{1.85E-3}$$^-$   & 3.65992$_{1.13E-3}$$^-$   & 3.65973$_{1.08E-3}$$^-$   & \cellcolor[rgb]{ .682,  .667,  .667}\textbf{3.66314 }$_\textbf{3.08E-4}$ \\
    UF2   & 3.63909$_{1.59E-2}$$^-$   & 3.64864$_{7.83E-3}$$^-$   & 3.65107$_{8.39E-3}$$^-$   & 3.65291$_{8.85E-3}$$^\sim$   & \cellcolor[rgb]{ .682,  .667,  .667}\textbf{3.65313 }$_\textbf{1.10E-2}$ \\
    UF3   & 3.62601$_{6.92E-2}$$^-$   & 3.64753$_{1.74E-2}$$^-$   & 3.65765$_{5.71E-3}$$^\sim$   & 3.65162$_{1.51E-2}$$^-$   & \cellcolor[rgb]{ .682,  .667,  .667}\textbf{3.65836 }$_\textbf{5.60E-3}$ \\
    UF4   & 3.14765$_{1.59E-2}$$^-$   & 3.18129$_{1.19E-2}$$^-$   & 3.17597$_{1.03E-2}$$^-$  & 3.17725$_{1.27E-2}$$^-$   & \cellcolor[rgb]{ .682,  .667,  .667}\textbf{3.26432 }$_\textbf{4.40E-3}$ \\
    UF5   & 2.56553$_{2.35E-1}$$^-$   & 2.71102$_{2.07E-1}$$^-$   & 2.90646$_{9.60E-2}$$^-$   & 2.92076$_{1.01E-1}$$^-$   & \cellcolor[rgb]{ .682,  .667,  .667}\textbf{3.07248 }$_\textbf{1.62E-1}$ \\
    UF6   & 2.90857$_{3.27E-1}$$^-$   & 2.84999$_{4.43E-1}$$^-$   & \cellcolor[rgb]{ .682,  .667,  .667}\textbf{3.17650 }$_\textbf{6.75E-2}$$^+$  & 3.14163$_{1.53E-1}$$^+$   & 3.02291$_{2.11E-1}$  \\
    UF7   & 3.46933$_{4.88E-2}$$^-$   & 3.49045$_{2.29E-3}$$^-$   & 3.49274$_{2.55E-3}$$^-$   & 3.49232$_{2.01E-3}$$^-$   & \cellcolor[rgb]{ .682,  .667,  .667}\textbf{3.49626 }$_\textbf{2.46E-4}$ \\
    UF8   & 7.18889$_{2.24E-2}$$^-$   & 7.17681$_{3.06E-2}$$^-$   & 7.19519$_{2.28E-2}$$^-$   & 7.19303$_{2.26E-2}$$^-$   & \cellcolor[rgb]{ .682,  .667,  .667}\textbf{7.27682 }$_\textbf{3.68E-2}$ \\
    UF9   & 7.06634$_{4.78E-1}$$^-$   & 7.28091$_{3.52E-1}$$^-$   & 7.55889$_{1.51E-1}$$^-$   & 7.53802$_{2.51E-1}$$^-$   & \cellcolor[rgb]{ .682,  .667,  .667}\textbf{7.59815 }$_\textbf{1.15E-1}$ \\
    UF10  & 3.49981$_{3.94E-1}$$^-$   & 2.42976$_{4.40E-1}$$^-$   & 0.89670$_{5.48E-1}$$^-$   & 0.56221$_{5.53E-1}$$^-$   & \cellcolor[rgb]{ .682,  .667,  .667}\textbf{4.49922 }$_\textbf{5.35E-1}$ \\
    WFG1  & 5.09320$_{1.42E-1}$$^-$   & 5.16428$_{6.33E-2}$$^-$   & 5.19093$_{9.00E-2}$$^-$   & 5.17729$_{9.63E-2}$$^-$   & \cellcolor[rgb]{ .682,  .667,  .667}\textbf{5.58013 }$_\textbf{2.70E-2}$ \\
    WFG2  & 10.02039$_{6.70E-1}$$^-$   & 10.50875$_{4.29E-1}$$^-$   & 10.37559$_{4.91E-1}$$^-$   & 10.46272$_{4.22E-1}$$^-$   & \cellcolor[rgb]{ .682,  .667,  .667}\textbf{10.58567 }$_\textbf{3.96E-1}$ \\
    WFG3  & 10.89220$_{1.19E-1}$$^-$   & 10.92397$_{8.95E-2}$$^-$   & 10.93098$_{8.32E-2}$$^\sim$   & 10.93077$_{8.34E-2}$$^\sim$   & \cellcolor[rgb]{ .682,  .667,  .667}\textbf{10.95286 }$_\textbf{2.00E-6}$ \\
    WFG4  & 8.40620$_{5.17E-2}$$^-$   & 8.41161$_{3.88E-2}$$^-$   & 8.51377$_{4.25E-2}$$^-$   & 8.49385$_{6.52E-2}$$^-$   & \cellcolor[rgb]{ .682,  .667,  .667}\textbf{8.53702 }$_\textbf{2.97E-2}$ \\
    WFG5  & 8.06233$_{1.18E-1}$$^-$   & 8.09696$_{1.20E-1}$$^-$   & 8.19163$_{8.07E-2}$$^-$   & 8.18601$_{8.67E-2}$$^-$   & \cellcolor[rgb]{ .682,  .667,  .667}\textbf{8.35099 }$_\textbf{4.20E-2}$ \\
    WFG6  & 8.22110$_{1.52E-1}$$^-$   & 8.39915$_{2.09E-1}$$^-$   & 8.37701$_{1.10E-1}$$^-$   & 8.34767$_{1.01E-1}$$^-$   & \cellcolor[rgb]{ .682,  .667,  .667}\textbf{8.49870 }$_\textbf{1.45E-1}$ \\
    WFG7  & 8.63700$_{1.25E-1}$$^-$   & 8.65283$_{9.48E-2}$$^-$   & 8.64190$_{1.24E-1}$$^\sim$   & 8.63866$_{1.25E-1}$$^\sim$   & \cellcolor[rgb]{ .682,  .667,  .667}\textbf{8.67611 }$_\textbf{3.00E-6}$ \\
    WFG8  & 8.59288$_{1.77E-1}$$^-$   & 8.64452$_{1.20E-1}$$^-$   & 8.66164$_{7.84E-2}$$^\sim$   & 8.65935$_{9.16E-2}$$^\sim$   & \cellcolor[rgb]{ .682,  .667,  .667}\textbf{8.67610 }$_\textbf{9.00E-6}$ \\
    WFG9  & 8.17868$_{5.21E-2}$$^-$   & 8.19778$_{4.88E-2}$$^-$   & 8.18257$_{7.02E-2}$$^-$   & 8.19077$_{4.86E-2}$$^-$   & \cellcolor[rgb]{ .682,  .667,  .667}\textbf{8.30853 }$_\textbf{3.39E-2}$ \\
     -/$\sim$/+ & 19/0/0 & 19/0/0 & 14/4/1 & 14/4/1& \\
    \bottomrule
    \end{tabular}}%
  \label{tab:HV}%
\end{table}%

\begin{figure}[t]
\centering
\includegraphics[width=.2\linewidth]{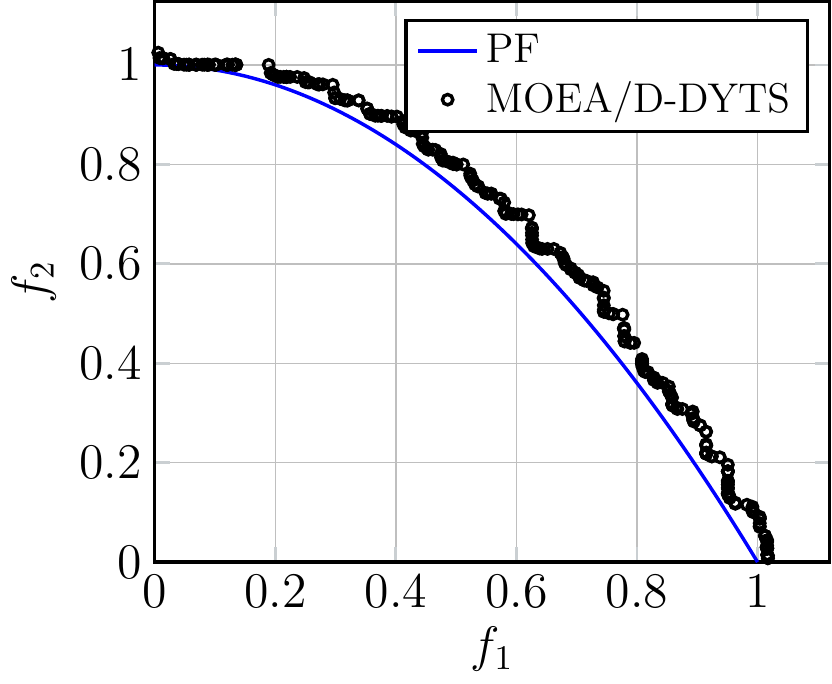}
\includegraphics[width=.2\linewidth]{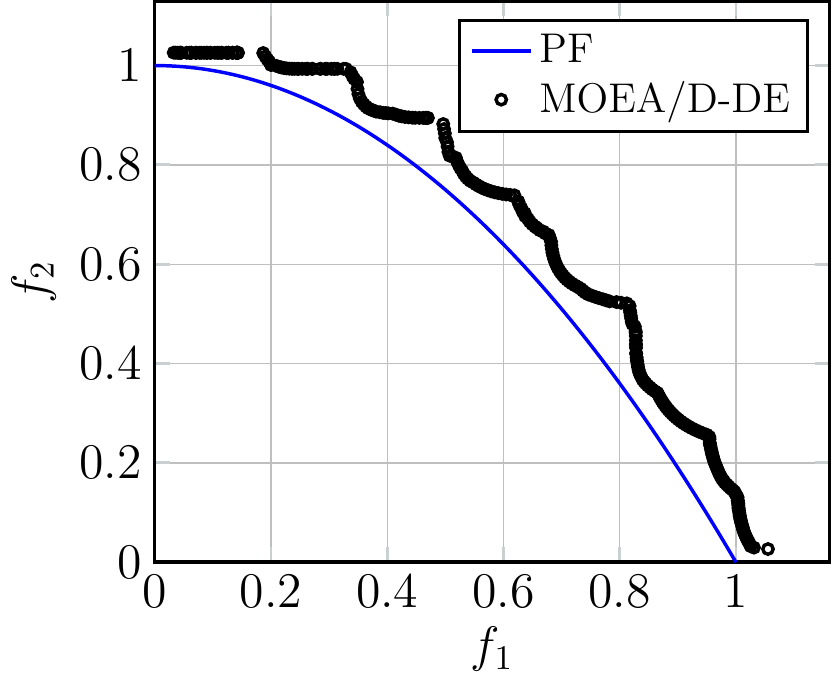}
\includegraphics[width=.2\linewidth]{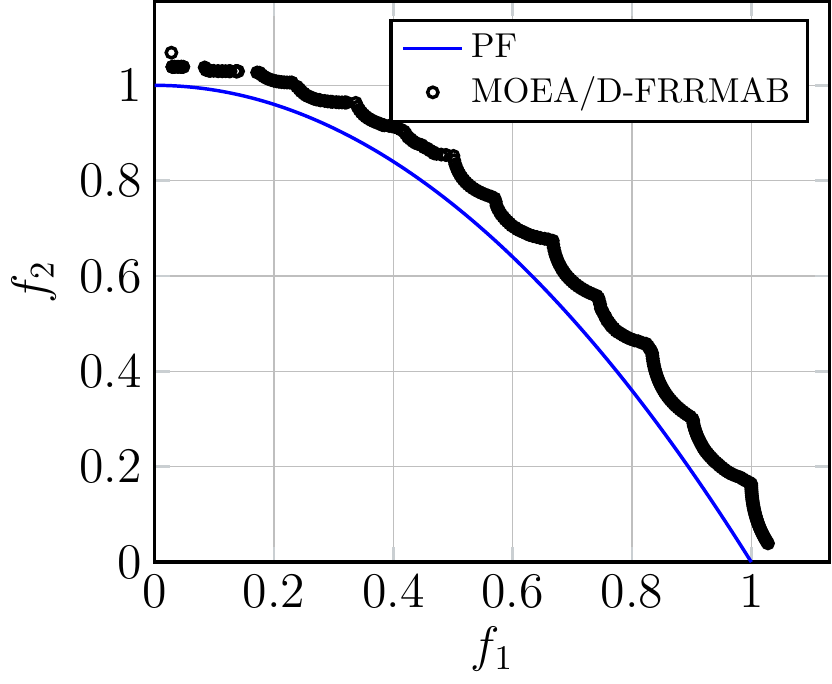}
\includegraphics[width=.2\linewidth]{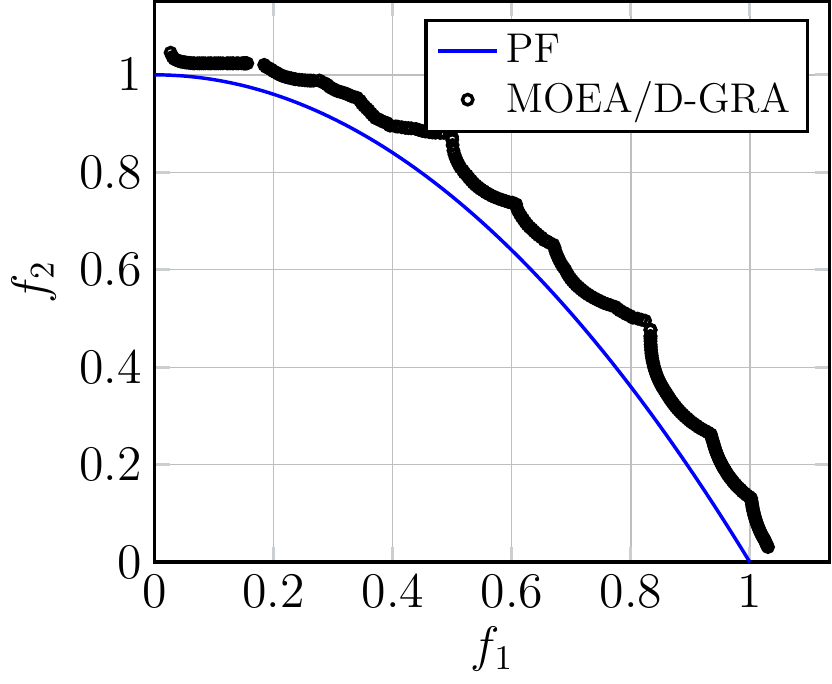}
\includegraphics[width=.2\linewidth]{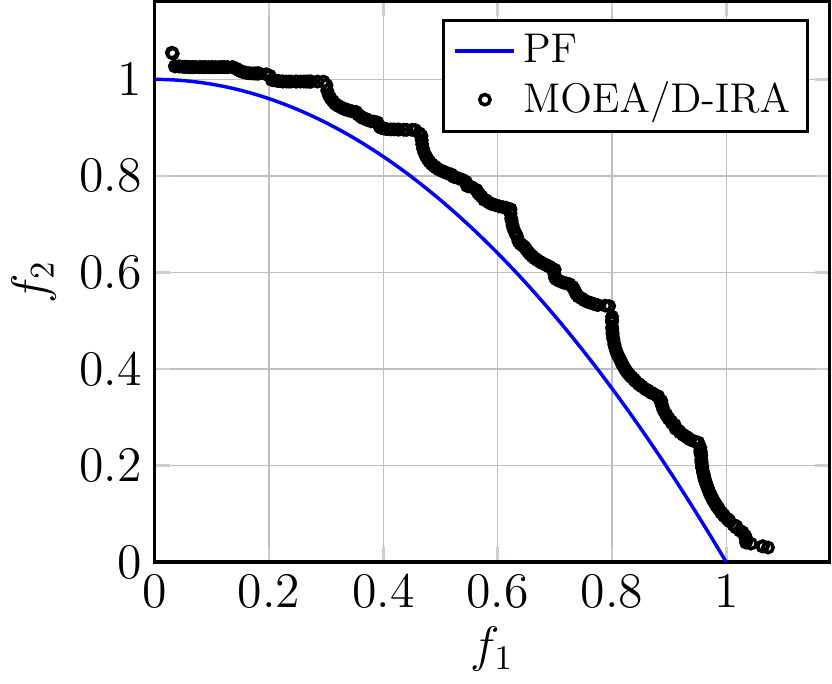}
\caption{Non-dominated solutions obtained by five algorithms on UF4 with the best IGD value.}
\label{UF4best}
\end{figure}

\begin{figure}[t]
\centering
\includegraphics[width=.2\linewidth]{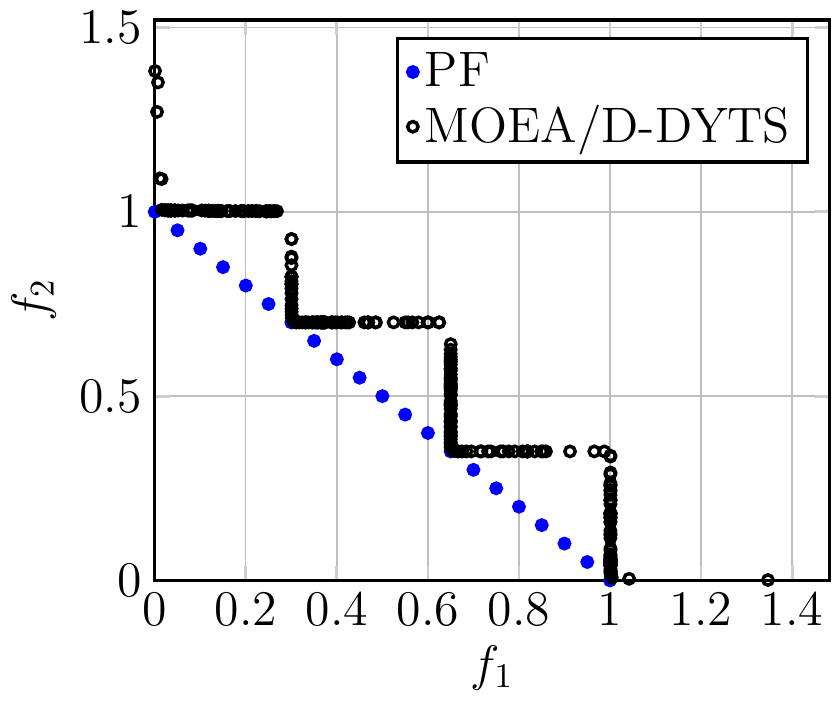}
\includegraphics[width=.2\linewidth]{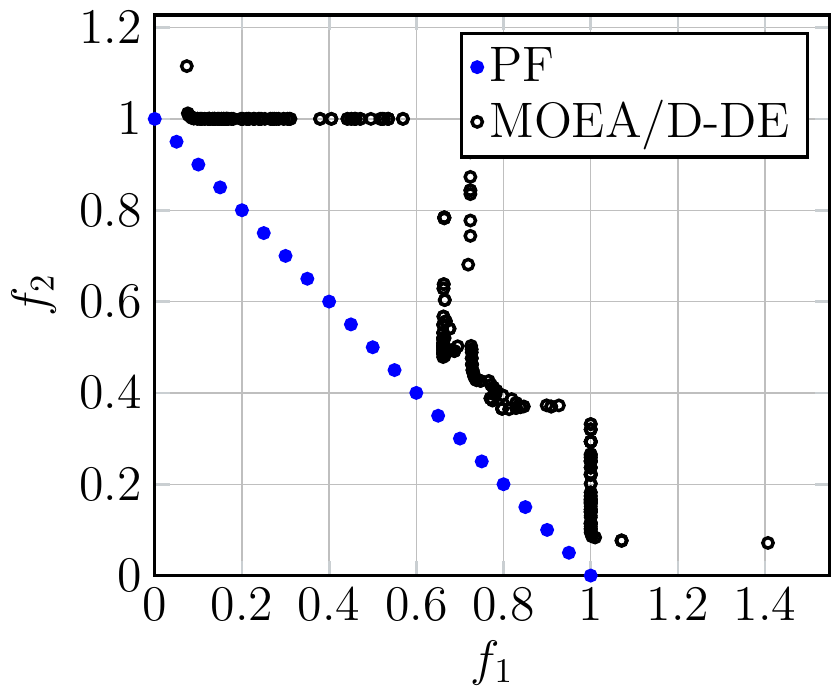}
\includegraphics[width=.2\linewidth]{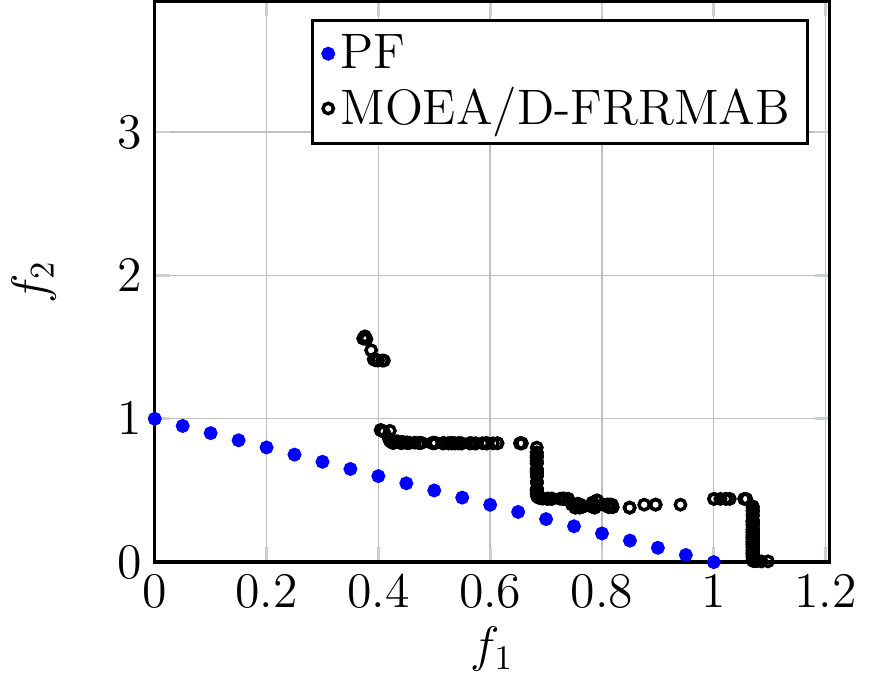}
\includegraphics[width=.2\linewidth]{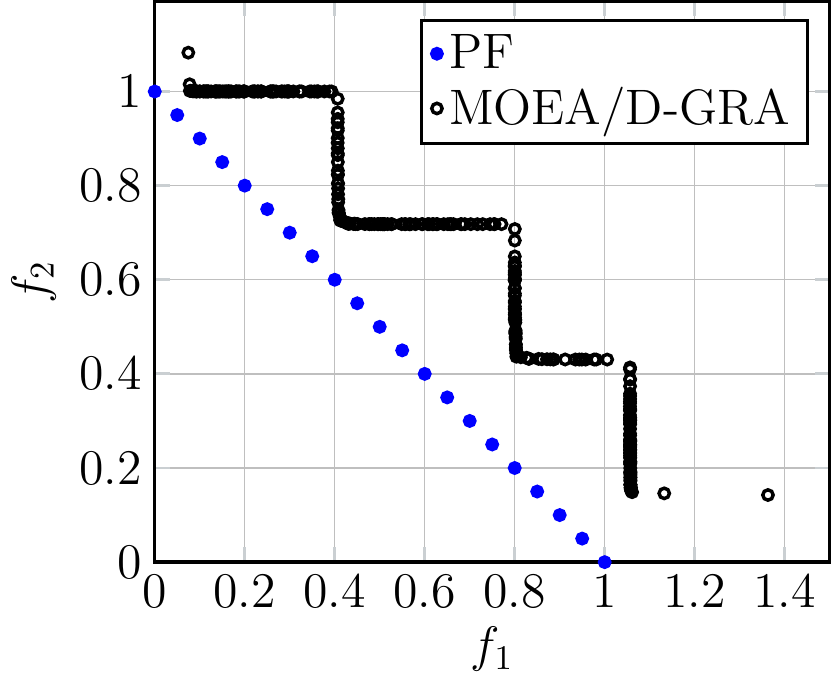}
\includegraphics[width=.2\linewidth]{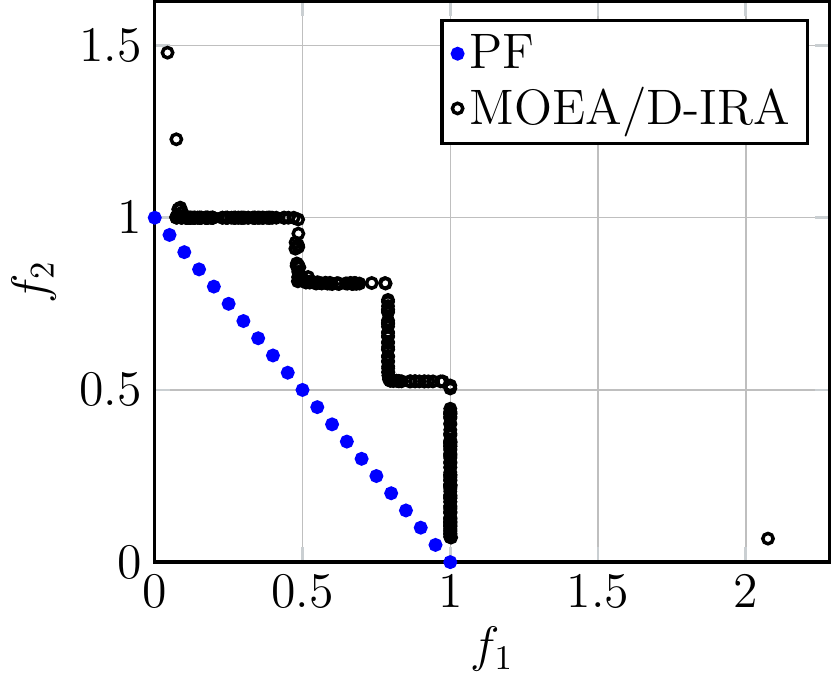}
\caption{Non-dominated solutions obtained by five algorithms on UF5 with the best IGD value.}
\label{UF5best}
\end{figure}

\begin{figure}[t]
\centering
\includegraphics[width=.2\linewidth]{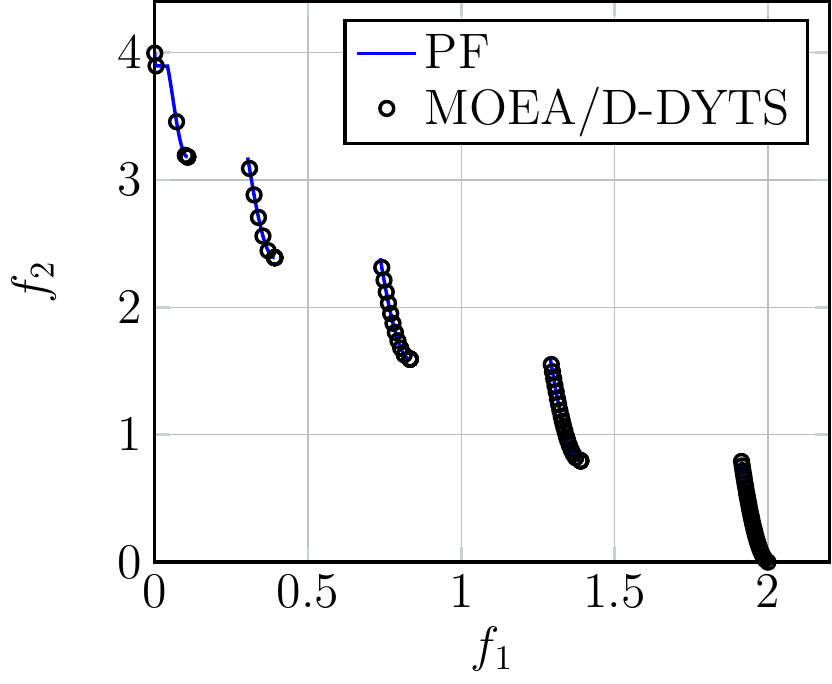}
\includegraphics[width=.2\linewidth]{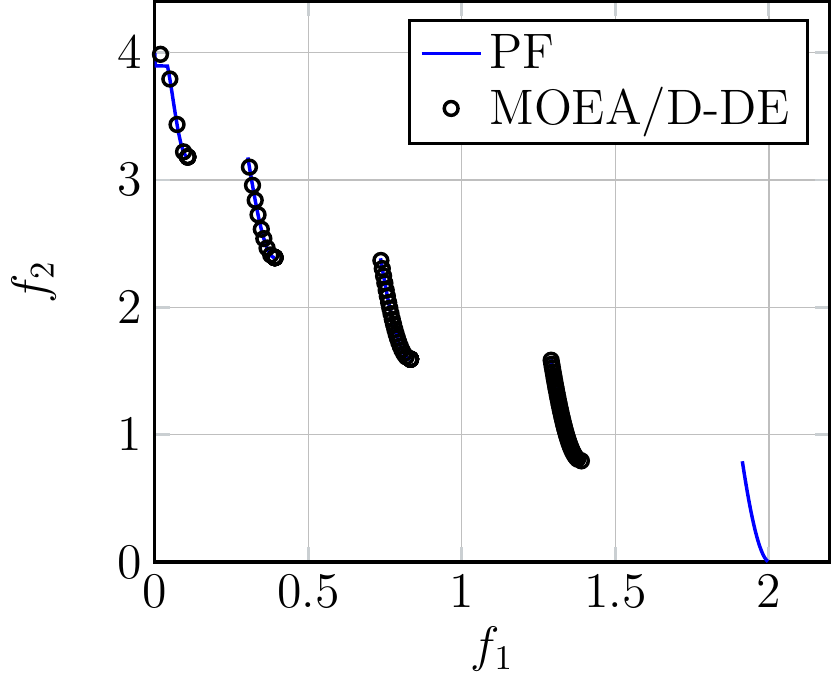}
\includegraphics[width=.2\linewidth]{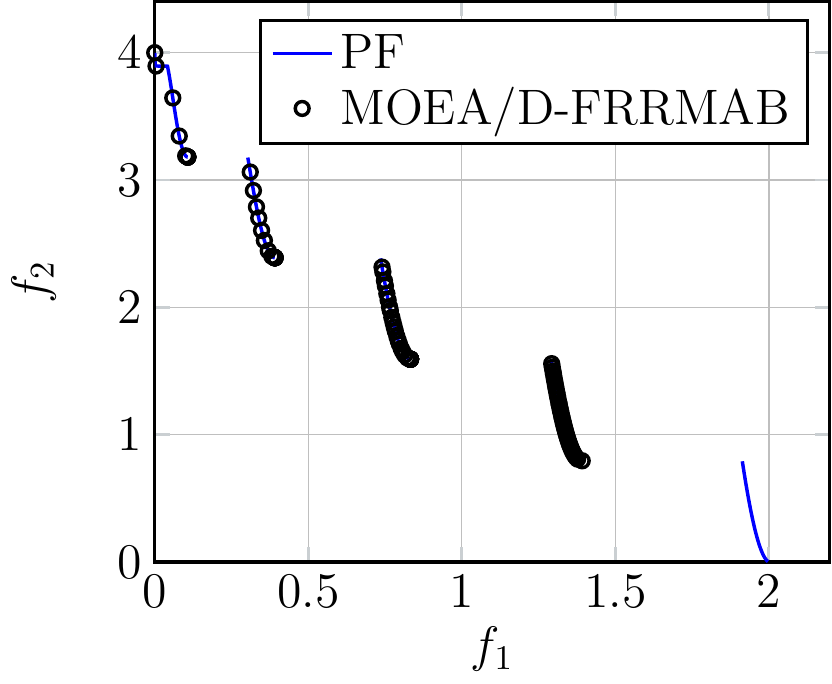}
\includegraphics[width=.2\linewidth]{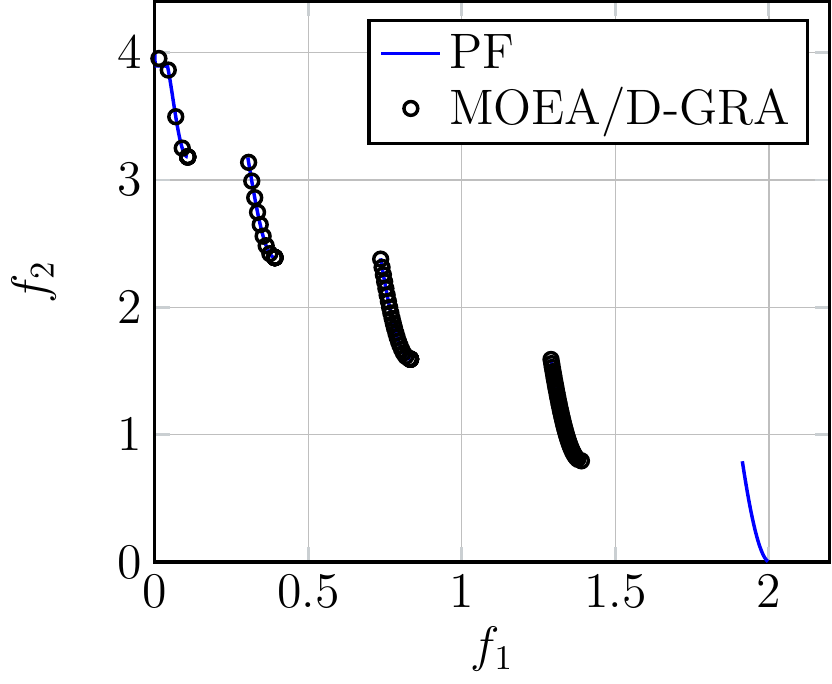}
\includegraphics[width=.2\linewidth]{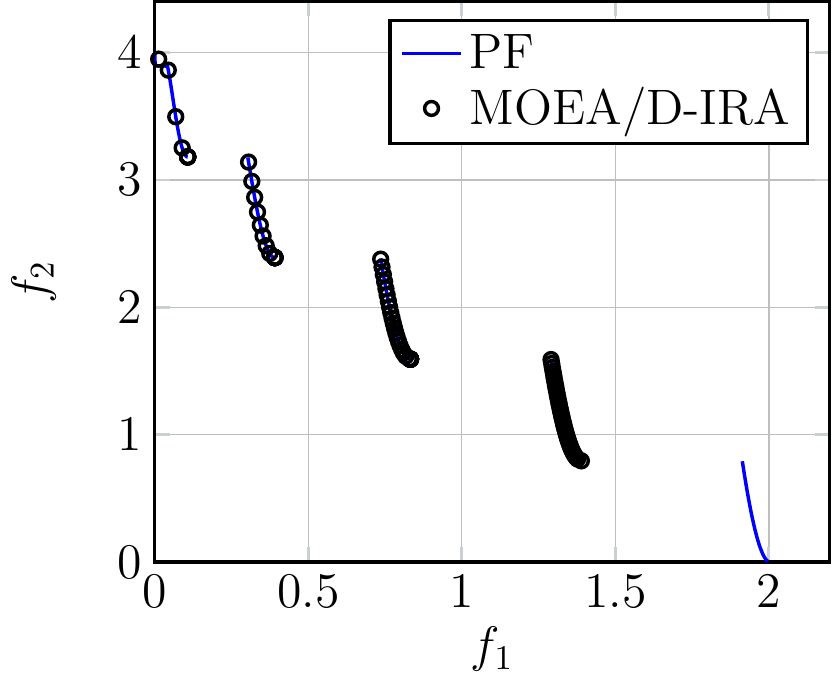}
\caption{Non-dominated solutions obtained by five algorithms on WFG2 with the best IGD value.}\label{WFG2best}
\end{figure}


From Tables \ref{tab:IGD} and \ref{tab:HV} we clearly see that MOEA/D-DYTS is the best algorithm. With respect to the IGD values, MOEA/D-DYTS performs best on UF1, UF3 to UF5, UF7, UF8, UF10 and all WFG test instances. Considering the HV values, Table \ref{tab:HV} gives the similar result to Table \ref{tab:IGD}. In total, it has obtained better results in 16 out of 19 performance comparisons for IGD and 18 out of 19 for HV. In the following paragraphs, we will give a gentle discussion over these results. 

UF1 to UF3 and UF7 are relatively simple test problems among UF benchmark problems, on which all the algorithms do not have too much difficulty to converge to the global PFs as shown in Figs. 1 to 3 and Fig. 7 in the supplementary document\footnote{https://cola-laboratory.github.io/supplementary/dyts-supp.pdf}. However, it is interesting to observe that MOEA/D-GRA achieves better convergency on the tail of the PF of UF2 on which our proposed algorithm takes a fall. On UF4 only MOEA/D-DYTS can find some solutions on the PF whereas the solutions found by other four algorithms are away from the PF. The PF of UF5 consists of 21 points which is challenging for EAs to converge. As shown in Fig.~\ref{UF5best}, the solutions obtained by MOEA/D-DYTS are much closer to the PF than the other algorithms on UF5. The similar results can also be observed on UF10. For other two three-objective UF test problems, all the algorithms can find most of the solutions of PF on UF8. Note that MOEA/D-DYTS is beaten by MOEA/D-GRA and MOEA/D-IRA on UF9 test problem which has two disconnected parts of PS. This may imply that MOEA/D-DYTS do not have enough search ability in decision space.

For WFG test problems, MOEA/D-DYTS has completely won all the 9 test instances. As observed from Tables \ref{tab:IGD} and \ref{tab:HV}, MOEA/D-DYTS perform significantly better than other four algorithms on WFG1. Regarding WFG2, it is a discontinuous problem whose PF is five disconnected segments. As shown in Fig.~\ref{WFG2best}, no algorithm can find solutions on the last segment on WFG2 except for MOEA/D-DYTS. For WFG3 to WFG9, all the compared algorithms can converge to the true PFs. The Wilcoxon rank sum test shows that MOEA/D-DYTS is similar to MOEA/D-GRA and MOEA/D-IRA on WFG3, WFG7 and WFG8 according to the results of HV values. It is worth noting that our proposed algorithm MOEA/D-DYTS has best mean values on WFG3, WFG7 and WFG8 from 31 independent runs. This implies that MOEA/D-DYTS can achieve better stability than other algorithms in optimisation process. As shown in Figs. 14 to 16 and 19 in the supplementary document, the better diversity of solutions found by MOEA/D-DYTS on the head or the tail of the PF can be observed on WFG4 to WFG6 and WFG9. These results indicate that our proposed algorithm can be favourable among other four state-of-the-art MOEA/D variants.


\section{Conclusion}
\label{sec:conclusion}

This paper proposes a new AOS paradigm for MOEA/D that is able to autonomously select the appropriate reproduction operator for the next step. Specifically, the dynamic Thompson sampling is served as the foundation for AOS. Different from the vanilla Thompson sampling and bandit learning model, the dynamic Thompson sampling is able to track the search dynamics under a non-stationary environment. From our empirical results, we have witnessed the superiority of our proposed MOEA/D-DYTS over four state-of-the-art MOEA/D variants on 19 test problems.

AOS is an attractive paradigm to equip EAs with intelligence to autonomously adapt their search behaviour according to the current search landscapes. In addition to the bandit model considered in this paper, it is also interesting to look into methods from reinforcement learning or automatic control domain for new decision-making paradigm. Furthermore, the credit assignment of an application of an operator is important to gear the AOS. More sophisticated methods are worthwhile being considered in future.

\section*{Acknowledgment}
K. Li was supported by UKRI Future Leaders Fellowship (Grant No. MR/S017062/1) and Royal Society (Grant No. IEC/NSFC/170243).

\bibliographystyle{IEEEtran}
\bibliography{IEEEabrv,mybib}

\end{document}